\documentclass[runningheads]{llncs}
\usepackage{graphicx}
\usepackage{amsmath,amssymb} 

\usepackage{algorithm}
\usepackage[noend]{algpseudocode}
\algnewcommand\algorithmicinput{\textbf{Input:}}
\algnewcommand\Input{\item[\algorithmicinput]}

\algnewcommand\algorithmicoutput{\textbf{Output:}}
\algnewcommand\Output{\item[\algorithmicoutput]}

\usepackage{color}
\usepackage{subcaption}
\usepackage{wrapfig,lipsum,booktabs}

\usepackage{capt-of}

\usepackage{multirow}

\usepackage{float}
\usepackage{dblfloatfix}
\usepackage{array}
\newcolumntype{P}[1]{>{\centering\arraybackslash}p{#1}}

\begin{document}

\title{Adaptively Denoising Proposal Collection for Weakly Supervised Object Localization} 

\titlerunning{Weakly Supervised Object Localization}

\author{Wenju Xu\and
		Yuanwei Wu\and
		Wenchi Ma\and
		Guanghui Wang}
%
\authorrunning{Wenju \it{et al.}}
%

\institute{School of Engineering, University of Kansas, Lawrence, KS, USA 66045}

\maketitle
\begin{abstract}
In this paper, we address the problem of weakly supervised object localization (WSL), which trains a detection network on the dataset with only image-level annotations. The proposed approach is built on the observation that the proposal set from the training dataset is a collection of background, object parts, and objects. Several strategies are taken to adaptively eliminate the noisy proposals and generate pseudo object-level annotations for the weakly labeled dataset. A multiple instance learning (MIL) algorithm enhanced by mask-out strategy is adopted to collect the class-specific object proposals, which are then utilized to adapt a pre-trained classification network to a detection network. In addition, the detection results from the detection network are re-weighted by jointly considering the detection scores and the overlap ratio of proposals in a proposal subset optimization framework. The optimal proposals work as object-level labels that enable a pseudo-strongly supervised dataset for training the detection network. Consequently, we establish a fully adaptive detection network. Extensive evaluations on the PASCAL VOC 2007 and 2012 datasets demonstrate a significant improvement compared with the state-of-the-art methods.
\end{abstract}




\section{Introduction}
\label{sec:intro}

Object detection, which attempts to place a tight bounding box around every object of a given image, is an important problem for image understanding. This problem has been extensively studied in recent years \cite{bergamo2014self,Alpher02,Alpher03,Authors14b,Authors14,song2014learning}, and the state-of-the-art detection performance promotes a variety of applications, including human pose estimation \cite{toshev2014deeppose} and crowd counting \cite{zhang2015cross}.
One key step for object detection is to learn a distinctive representation of the objects from a large quantity of labeled data. Most existing methods rely on object-level labeled dataset \cite{deng2009imagenet} so that their models learn visual features from those specified regions. However, data annotation is an exhaustive and error prone work. In order to reduce the annotation cost, a common strategy is to learn the detector in a weakly supervised manner that only binary image-level labels representing the overall presence or absence of an object category are added to the images for training.

Multiple instance learning (MIL) \cite{cinbis2014multi,deselaers2012weakly} is an intensively used strategy in dealing with the task of weakly supervised object localization (WSL). It selects object regions of interest (proposals) from the positive images that contains the object, and learns an appearance model of the object from the features in the selected regions. This method has the tendency to get stuck in local optima. Therefore, a re-localization and re-training strategy is typically taken to push the solution close to the global optima. Pentina \textit{et al.} \cite{pentina2015curriculum} forms a curriculum learning strategy to feed the training process from easy images with big objects to hard images with many small objects. Shi \textit{et al.} \cite{Authors14} propose a strategy that re-weights the proposals' scores according to the consistence between the proposal size and the estimated object size. Even though these strategies attempt to improve the MIL, finding positive image bags containing certain class object for MIL classifier, in some senses, depends on guessing and it is possible to take a negative bag as a positive one. It is also difficult to get tight bounding boxes exactly containing the objects. These drawbacks require strategies that adaptively refine the estimated bounding boxes to tightly contain the objects.

Another line of this research is based on convolutional neural networks (CNNs) \cite {jie2017deep,tang2017multiple} that are capable of learning generic visual features generalized to many tasks. Methods in this category are inspired by the facts that, without location annotation, a pre-trained image classification CNN learns representative information of objects and object parts. Many efforts leverage CNN to extract discriminative appearance features, then train a MIL appearance model for object detection \cite{wu2015deep}. Recent efforts \cite{Alpher03,Authors14b} achieve significant performance improvement with proposed end-to-end methods, which adopt a pre-trained classification network to mine location information and transfer the problem from weakly supervised object localization to psudo-strongly supervised object detection. However, generating instance-level labels from the image-level labels is nontrivial since the  objects from the same category may appear with different shapes and background. A pre-trained classifier makes predictions on salient features. The extracted appearance features represent object parts, which lack information on the instance as a whole. Moreover, it is different to determine the size of bounding boxes that exactly contain the objects in the feature-level searching. As a result, the obtained instance-level labels are inexact.
\begin{figure*}
	\begin{minipage}[b]{1.0\linewidth}
		\centering
		\centerline{\includegraphics[width=12cm]{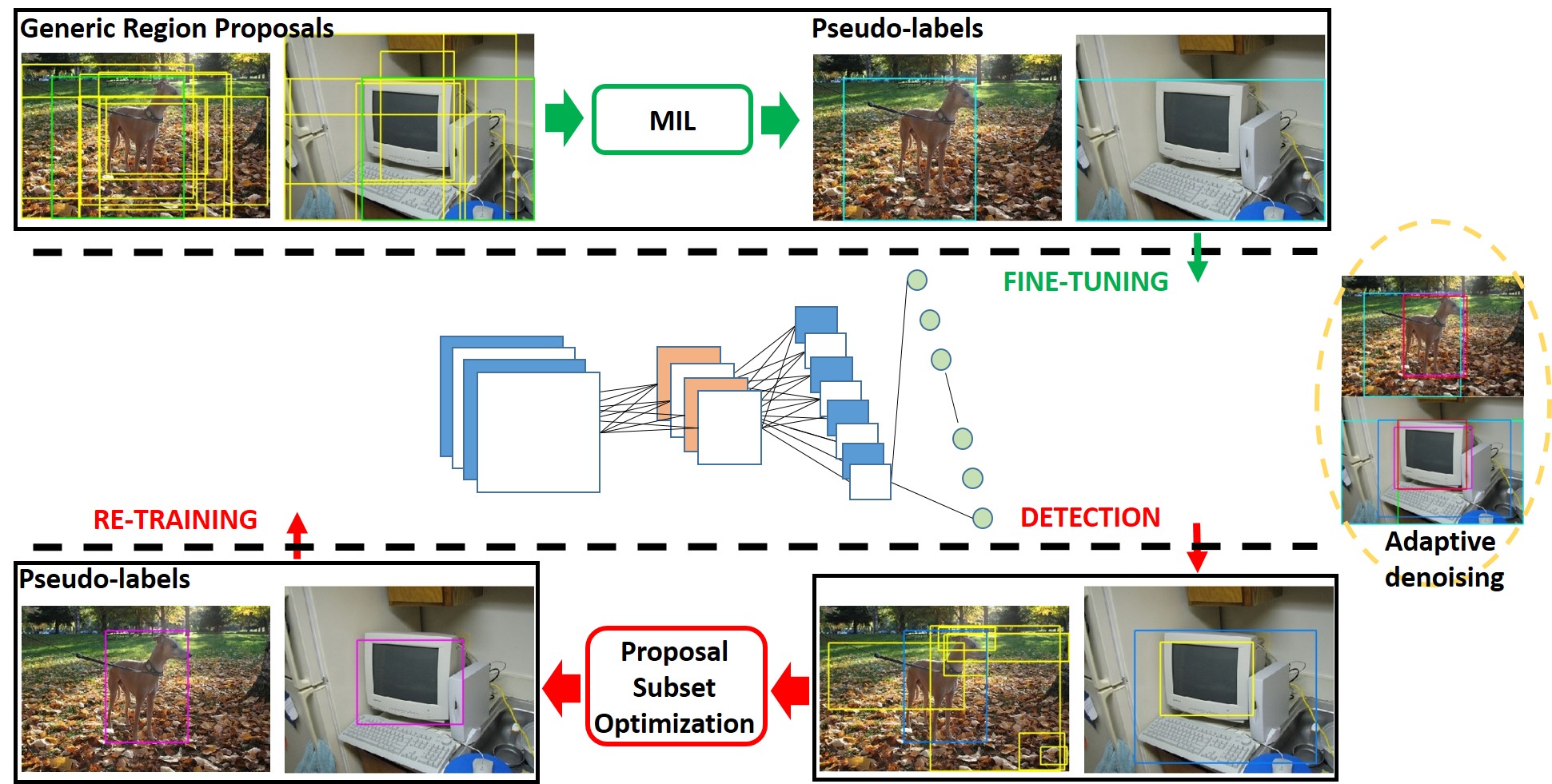}}
	\end{minipage}
	\caption{Overview of our method. We use mask-out strategy to collect the generic region proposals and take the MIL to generate pseudo labeled training set. This dataset is then fed to a WSL loop, so that the object detector is re-trained  progressively.  We also take the re-localization \cite{zhang2015improving,zhu2015segdeepm} step by re-weighting object  proposals according to the detection scores and the overlap of the proposals. Bounding boxes (in yellow) represent the confident proposals; while the bounding box in other colors in each block represents the highest confident proposal.}
	\label{fig_1}
\end{figure*}
In this paper, we propose a new framework based on two observations: (i) The proposals are a collection of background, object parts, and objects; and (ii) it is hard to train object detectors directly under weakly labeled dataset due to the substantial amount of noise in the object proposal collection and the size variation of the objects. Our method integrates several strategies to adaptively eliminate the noise in the object proposal collection. We take an enhanced MIL algorithm, which is proceeded by a mask-out strategy to mine the proposal collection and fine-tune a pre-trained classification network through re-weighting and re-training, which exploits proposal subset optimization \cite{zhang2016unconstrained} to further re-weight the detection results.

Our re-weighting and re-training strategy aims at determining the optimal proposals automatically.
To this end, we take a subset optimization method to select object proposals. It is based on both
the detection scores from the pre-trained detection network and the overlaps between the candidate bounding boxes. This strategy puts higher weights on proposals that have large overlap area with others. Specially, We reweight those object proposals with high detection scores
according to how much the bounding box overlaps with other bounding boxes. Iteratively, we utilize this subset optimization method to improve the re-localization step. 

\begin{figure*}
	
	\begin{minipage}[b]{1.0\linewidth}
		\centering
		\centering{\includegraphics[width=12cm]{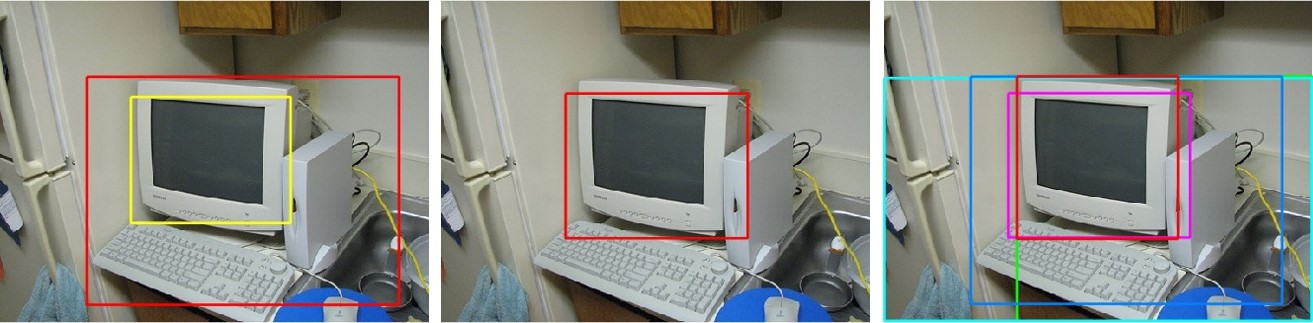}}
	\end{minipage}
	\caption{Detection results from NMS (red line in left) and subset optimization (center). Bounding boxes (BB) (right) represent the highest confident proposals got from different steps (blue BB: CNN, green BB: maskout, pink BB: re-train, cyan BB: MIL). We compare the detection results by bounding boxes in different colors, which shows our re-training strategy is able to get the denoising proposals by re-weighting object proposals according to the detection scores and the overlap ratios of the proposals.}
	\label{fig_5}
\end{figure*}

This re-weighting scheme reduces the
uncertainty in the proposal distribution, making the re-weighting step more
likely to pick a proposal correctly covering the object. Fig.~\ref{fig_5} shows an example
of how the subset optimization changes the proposal score induced by the
current object detector, leading to a more accurate localization.

Our contributions are as follows: 
(i) We propose a novel work flow to collect confident proposals, which integrates the mask-out strategy, MIL, and subset proposal optimization. The MIL model is trained on the selected proposals of mask-out strategy and mines confident proposals to reduce the background clutters and potential confusion from similar objects cases. The subset proposal optimization further refines the proposals by re-scoring the bounding box;
(ii) Following the idea of re-localization and re-training, the candidate proposals are refined based on both the detection scores and the overlap ratios between the proposals. We then iteratively adapt a pre-trained classification network to a detection network with those quality enhanced proposals. This is a new pipeline for improving object proposals;
And (iii) detailed evaluations on the PASCAL VOC 2007 and 2012 datasets \cite{everingham2015pascal} demonstrate that our weakly supervised object detection with adaptively denoised proposal collection performs favorably against the state-of-the-art methods. The proposed model and trained parameters will be available on the authors website.

	
	%
	%
	%

\section{Related work}
Extracting meaningful information from the environment is a challenging task \cite{xu2018direct,xu2016direct,zhou2014smart}. In recent years, deep neural networks are becoming more and more popular for knowledge discovering in many computer vision tasks, such as object recognition \cite{xu2019towards,zhang2018bpgrad}, object detection \cite{liu2012pedestrian,kong2017cancer}, visual question answering \cite{yu2018beyond}, pose estimateion \cite{hong2015multimodal}, image synthesis \cite{xu2019,XU2019570,XU2019195}, face recognition \cite{cen2019dictionary}, and depth estimation \cite{he2018learning}. Object detection is the task of recognizing and localizing the objects in the images with the deep model trained on labelled ground truth \cite{ma2018mdcn}. However, labelling the images with bounding box for each object is a nontrivial work. In the scenario of weakly supervised localization, the training
images are known to containing instances of a certain object class but their locations. There is no ground truth bounding box available for each object in the training dataset. The task is both to localize the objects (estimate the bounding boxes tightly containing the instances) and to classify the objects. What we have are the image-level annotations which are weak supervision for localizing the objects. To train a detection network with image-level supervision, we need first to localize objects in all the images of the training dataset based on image-level annotations, and then use the localization results to train a detector for the test set. The WSL problem is often handled with multiple
instance learning (MIL) \cite{Alpher02,Alpher03,cinbis2017weakly,song2014learning}, where the images are treated as bags of
object proposals \cite{zitnick2014edge} (which are bounding boxes estimated to localize the objects). A negative image dose not contain instances of certain category. A positive image contains at least one positive instance, mixed in with
a majority of negative ones. The goal is to find the true positive instances from
which to learn a classifier for proposal classification. 

Previous works achieve significant improvement by exploring ways to enhance the MIL. Siva \textit{et al.} \cite{siva2012defence} propose an
effective negative mining approach combined with discriminative saliency measures.
Song \textit{et al.} \cite{song2014learning} formulate an initialization strategy for WSL
as a discriminative submodular cover problem in a graph-based framework,
and develop a negative mining technique to increase robustness against incorrectly
localized boxes \cite{song2014weakly}. Bilen \textit{et al.} \cite{Alpher02,Alpher03} propose a relaxed version of MIL that
softly labels object instances instead of choosing the highest scoring ones.
They also propose a discriminative convex clustering algorithm to jointly learn
a discriminative object model and enforce the similarity of the localized object
regions.

As CNNs have turned out to be surprisingly effective in many vision tasks
including classification and detection, recent state-of-the-art WSL approaches
also build on CNN architectures \cite{Alpher03} or CNN features \cite{cinbis2017weakly}. Bilen \textit{et al.} \cite{Alpher03} modify a region-based CNN architecture \cite{girshick2015fast} and propose a CNN with
two streams, one focuses on recognition and the other one on localization, which
performs simultaneously region selection and classification. Similarly, Li \textit{et al.} \cite{Authors14b} use the MIL to obtain the initial detection results and propose a domain adaption method \cite{hoffman2014lsda} to fine-tune a classification network into a detection network with the initial detection results. The results show a performance improvement on the detection accuracy. Shi \textit{et al.} \cite{Authors14} attempt to score the proposals by the size and retrain the detection network with the re-weighted proposals according to an easy to hard order, based on the assumption that the proposals with bigger size provide more information to train the network than the those with smaller size. Our work is related to these CNN-based MIL approaches that perform WSL by end-to-end training
from image-level labels. In contrast to the above methods, however, we focus on
a CNN architecture that is re-trained in an order for detection accuracy improvement with denoised proposals. 

The concept of adaptive learning in an order was also studied in computer vision \cite{lapin2014learning,pentina2015curriculum}. These works focus on a key question: how to re-weight the proposals? Sharmanska \textit{et al.} \cite{sharmanska2013learning} employ
some privileged information to distinguish between easy and hard examples in
an image classification task. The privileged information are additional cues available
at training time, but not at test time. They employ several additional cues,
such as object bounding boxes \cite{guillaumin2012large}, image tags and rationales to define their concept
of easiness \cite{lapin2014learning}. Lai \textit{et al.} \cite{lai2017saliency} select highly confident object proposals under the guidance of class-specific saliency maps. Pentina \textit{et al.} \cite{pentina2015curriculum} consider learning the visual attributes of objects. Shi \textit{et al.} \cite{Authors14} propose a size estimator to re-weight the proposals based on the size of the instances in the image. They use curriculum learning in a WSL setting and propose object size
as an "easiness" measure. Shi \textit{et al.} \cite{shi2017weakly} consider the task of discovering object classes in an
unordered image collection. their model is initialized with regions of "stuff"
categories, and is then used to support discovering "thing" categories in unlabelled
images with the help of a fully supervised segmentation model. Bodla \textit{et al.} \cite{bodla2017improving} propose a soft method to select the bounding boxes. They decay the classification score of a box which has a high overlap with top-scoring boxes, rather than suppressing
it. Jie \textit{et al.} \cite{jie2017deep} explore the Fast RCNN model \cite{girshick2015fast} and propose a self-taught learning method for proposal selection. The most related work to ours is the very recent study \cite{tang2017multiple},
which designs an on-line classifier refinement pipeline to progressively locate the most discriminative region of an image. By contrast, we propose a novel work flow to adaptively refine the proposals, i.e., to iteratively collect a more confident subset of proposals. In addition, we take the re-training strategy to fine-tune the model with the denoised proposal subset.
The proposed work flow, by integrating several novel proposal mining strategies, makes it adaptable to a variety of weakly supervised object detection tasks.

\section{Adaptively Denoised Proposal Collection}
\label{sec:format}
The proposed weakly supervised object detection method is illustrated in Fig.~\ref{fig_1}. This model consists of three major components, namely confident proposal learning, object detector learning and proposal subset optimization. They are successively employed to adaptively refine the proposal collection. The remainder of this section discusses these three components in details.

\subsection{Confident Proposal Mining}
We consider the weakly supervised object localization problem as an adaptively proposal denoising procedure that gradually refines the proposal collection. At the end, we transfer the problem from the weakly supervised object localization to a pseudo-strongly supervised object detection. Based on a pre-trained CNN classification network and a MIL model, our work flow adaptively selects confident proposals other than those comprised of background or object parts from the candidate proposals generated by EdgeBoxes \cite{zitnick2014edge}.

Assisted by the classification network, we first utilize the mask-out strategy to collect object proposals. The idea of masking out the input of CNN has been previously explored in \cite{zeiler2014visualizing}, which replaces the pixel values of the proposals with fixed mean pixel values; and compares the classification scores of feeding the real image and its mask-out images into the classification network. Intuitively, if the mask-out image introduces a notable drop in the classification score for the $c_{th}$ class, the region can be considered as containing an object belonging to the $c_{th}$ class. Inspired by \cite{zeiler2014visualizing}, we apply the mask-out strategy to select the proposals containing a certain object. We denote the classification network as $f_c$ that maps an image to a confidence vector of $c_{th}$ classes. The confident proposals $B_c$ are selected by investigating the difference of classification score between the selected image $I(x)$ and its mask-out image $I(x/b)$. This is formulated as
\begin{align}
\label{eq1}
\begin{split}
B_c=arg\max_{b}(f_c(I(x)) - f_c(I(x/b))) 
\end{split}
\end{align}
where $b$ represents the masked-out region. To select confident proposals, we first set a threshold on the classification score. The region $b$ is considered discriminative for $c_{th}$ class based on two aspects: the score of classifying the image $I(x)$ to the $c_{th}$ class is beyond the threshold and the classification score drop between the image and corresponding mask-out images is maximum. 

Once the proposals are obtained by applying the mask-out strategy, we separately learn one MIL model for each category. Taking the purified proposals selected by the mast-out strategy as training dataset makes the basic MIL initialized from a higher baseline, which not only stabilizes the training process, but also reduces the time for training \cite{Authors14b}. In the MIL model, each instance is described by a feature vector. More specifically, each feature vector is regarded as an instance and each image is represented by a bag of instances. For instance, the training image $x_i$ is considered as a bag of proposals with pseudo strong labels $y_i \in \{-1, 1\}$ indicating whether the bag contains an instance in the specific category. A bag is considered to be negative if there is no instances or all its instances are not in that category, while it is positive if there is at least one of its instances in that category. Given feature representation $\phi(x_i,z)$, we iteratively train the MIL model with the objective written as 
\begin{align}
\begin{split}
\min_{w \in R}\frac{1}{2}||w||_2^2 - \sum_{i=1}^{n}log((y_i\max_{z\in \mathcal{Z}}w^T\phi(x_i,z) -\frac{1}{2})+\frac{1}{2})
\end{split}
\end{align}
where $w$ represents the parameters of the MIL model and $z$ is called the "latent variable" chosen from the set $Z$, which is typically a set of bounding boxes. The top-scoring proposals given by the mask-out strategy are taken as positive samples for each category, which are used to train the MIL model. Among the initial bounding boxes, the set $Z$ contains all
possible candidate instances. Maximizing the objective function over $Z$ amounts to choosing a bounding box containing the whole object. The proposal, in this work, is represented by a 4096-dimensional feature vector from the second-last layer of the classification network. 

The top row in Fig.\ref{fig_1} demonstrates the idea of the confident proposal mining, which starts from the mask-out strategy and ends with the high confident output from MIL.

\subsection{ Proposal Subset Optimization}
Proposal selection based object detection method has one severely issue of overlapping among the bounding boxes that correspond to the same object. To select the best bounding box for each object, greedy non-maximum suppression (NMS) is widely employed as the latest strategy which selects the top-scoring bounding box $b_i$ and discards other bounding boxes $\mathcal{M}$ that have overlaps with the chosen one larger than a threshold $T$. Due to simplicity, this NMS mainly focus on the detection score $s_i$. By taking the Intersection over Union (IoU) as the measure of overlapping, this non-maximum suppression process can be described as

\begin{align}
\begin{split}
s_i =
\begin{cases}
s_i   ~~~~~~IoU(\mathcal{M},b_i)<T;\\
0 ~~~~~~~~ IoU(\mathcal{M},b_i)\geq T.\\
\end{cases}
\end{split}
\end{align}
However, there are no instance-level labels available for network training in the weakly supervised localization task, even the bounding boxes estimated with top score are tended to be noisy.
To overcome this issue, we propose a subset optimization scheme. It is realized by re-weighting the detection scores among the bounding boxes with high but noisy initial scores, where greedy NMS is not able to adjust the estimated bounding box accordingly. The proposed approach is similar to that described in \cite{zhang2016unconstrained}. However, we employ the method to solve the weakly supervised learning problem. The confident proposals with high detection scores are grouped into clusters by jointly considering the scores and the spatial overlaps between the proposals. The bounding box set is represented by $B=\{b_i:i=1:n\}$. We denote the group membership as $X=(x_i)^n_{i=1}$, where $x_i=j$ if $b_i$ belongs to a cluster $b_j$. Then one exemplary bounding box $o$ is selected from each cluster $B$ as the final output. This is formulated as finding the maximum a posterior (MAP) solution of the joint distribution $P(O,X|I;B,S)$, which tends to assign big value to the bounding boxes that have large overlap with more confident proposals. After taking the log of the posterior, the objective function becomes:
\begin{align}
\label{eq4}
\begin{split}
O=X^{\ast}=arg\max_{X}\sum_{i=1}^{n}\omega_i(x_i)
\end{split}
\end{align}
where $\omega_i(x_i=j)=logP(x_i=j|I)$

\begin{align}
\begin{split}
\mathrm{P(x_i=j|I)} =
\begin{cases}
Z^i_2 \lambda   ~~~~~~~~~~~~~~~~~~~~~~~~if~ j=0;\\
Z^i_2K(b_i,b_j)s_i~~~~~~~~ otherwise.\\
\end{cases}
\end{split}
\end{align}
$ K(b_i,b_j)$ is the window IoU used to measure the spatial overlap between $b_i$ and $b_j$, $S=\{s_i:i=1:n\}$ is the score set containing the detection scores of all the bounding boxes, and $Z^i_2$ is the normalization constant. Parameter $\beta$ and $\gamma$ control the penalty level. Note that our proposal subset optimization method takes both the scores and the overlapping into consideration since the detection scores in the weakly supervised task are not always reliable. 

The proposal subset optimization problem is defined as:
\begin{align}
\label{eq6}
\begin{split}
O^{\ast}=arg\max_{O}\beta\sum_{i\in \hat{O}}s_i-\gamma\sum_{i,j\in \hat{O}:i\neq j}K(b_i,b_j)
\end{split}
\end{align}

In this setting, we first maximize the objective function over $X$ according to Eq. (\ref{eq4}), which will select the cluster centers. Then, a greedy algorithm is used to choose a minimal number of bounding boxes as the outputs based on Eq. (\ref{eq6}). More details of the method can be found in \cite{zhang2016unconstrained}.

\subsection{Object Detector Learning}
In this step, we adapt the pre-trained classification network to an object detection network. This neural network is trained with the pseudo labeled proposals obtained from the proposal subset optimization strategy. We employ the re-weighting and re-training strategy for network adaption. The network parameters are fine-tuned for object localization, as illustrated in the bottom of the Fig.\ref{fig_1}. We organize it as adaptively refining the proposal subset, which is similar to the curriculum learning. However, we do not separate the training dataset into easy and hard parts. We start by running MIL, which is initialized with the results from mask-out strategy. This leads to a reasonable first detection model $A_1$. We move forward by running proposal subset optimization on the proposals subset with high detection scores, which produce a re-weighted proposal subset. The process then moves on to the second training iteration, where the training dataset consists of re-weighted proposals with more confident pseudo labels. As a result, the refined model $A_2$ will localize the objects better than $A_1$, as it is trained with better supervision in the re-training step. The process iteratively moves on to the next round, starting from the detection model $A_k$ and yielding a better one $A_{k+1}$. The whole training procedure is described in Algorithm \ref{algo}.

The selected results from each strategy are shown in Fig.~\ref{fig_5}. It is demonstrated that the bounding boxes selected by the fully adapted detection network exactly contain the objects, while the bounding boxes selected by mask-out strategy and MIL contain the object but with a large margin. By re-weighting the confident proposals according to the detection scores and the overlap of the proposals, the re-training strategy is able to generate more confident proposals.

\begin{algorithm}
	\caption{The training pipeline of the proposed algorithm.
	}\label{algo}
	\begin{algorithmic}[1]
		\Input 
		\Statex Images $x \in X$; $B=\{b_i:i=1:n\}$ candidate boxes; $S=\{s_i:i=1:n\}$ the corresponding scores; K, the refinement times; M, the network iteration times; $\theta_E$, network parameters. 	
		\Output
		\Statex A fully adaptive detection network.
		\State Classify the real images and the mask-out images with the classification network; select the top M proposals by Eq. (\ref{eq1}) as the initial proposal set $P_0 \gets \{x,s_0, B_0\}$.
		\State For each category, construct positive and negative bags within $S_0$; train the MIL model and collect the detection results from the trained MIL model as proposal set $P_1 \gets \{x,s_1, B_1\}$.
		\For {$k=1  ~\textbf{to}~  K-1$}
		\State Set $P \gets P_k$.				
		\For {$ m=1  ~\textbf{to}~  M-1$}
		\State \parbox[t]{\dimexpr\linewidth-\algorithmicindent}{Sample $P \to \{x, s, b\}$ as a minibatch.}
		\State Network forward propagation and get loss $\ell$.
		\State \parbox[t]{\dimexpr\linewidth-\algorithmicindent}{Network backward propagation, $\theta_E \stackrel{+}\leftarrow -\nabla_{\theta_E}(\ell) $.}
		\EndFor
		\State \textbf{end}
		\State \parbox[t]{\dimexpr\linewidth-\algorithmicindent}{Collect the detection results from the trained detection network, $P'_k  \gets \{x_k, s_k, B_k\}$.}
		\State 
		\parbox[t]{\dimexpr\linewidth-\algorithmicindent}{Choose the proposals with subset optimization by Eq. (\ref{eq6}); update the proposal set $P_{k+1} \gets \{x, s'_k, B'_k\}$.}
		\EndFor
		\State \textbf{end} 
	\end{algorithmic}
\end{algorithm}

\section{Experimental Evaluation} 
{\bf Dataset and settings:}
The proposed approach is extensively evaluated on two publicly available datasets: PASCAL VOC 2007 and 2012 datasets. Both of them have 20 classes of different images.
We employ both the AlexNet \cite{krizhevsky2012imagenet} and VGGNet
\cite{simonyan2014very} as our base CNN models, initialized with
parameters transferred from the classification network, which is pre-trained on the ImageNet dataset. As an initialization step for class-specific proposal mining, we use Edge Boxes \cite{zitnick2014edge} to generate 2,000 object proposals for each image. The mask-out strategy is first utilized to remove most of the noisy proposals and return top 50 confident proposals. These selected proposals work as the input for multiple instance learning. At the re-training stage, network is trained by employing the SGD solver with the learning rate of 0.0001 for 40k iterations.
\begin{figure*}[t!]
	\begin{minipage}[b]{1.0\linewidth}
		\centering
		\centerline{\includegraphics[height=4.25cm]{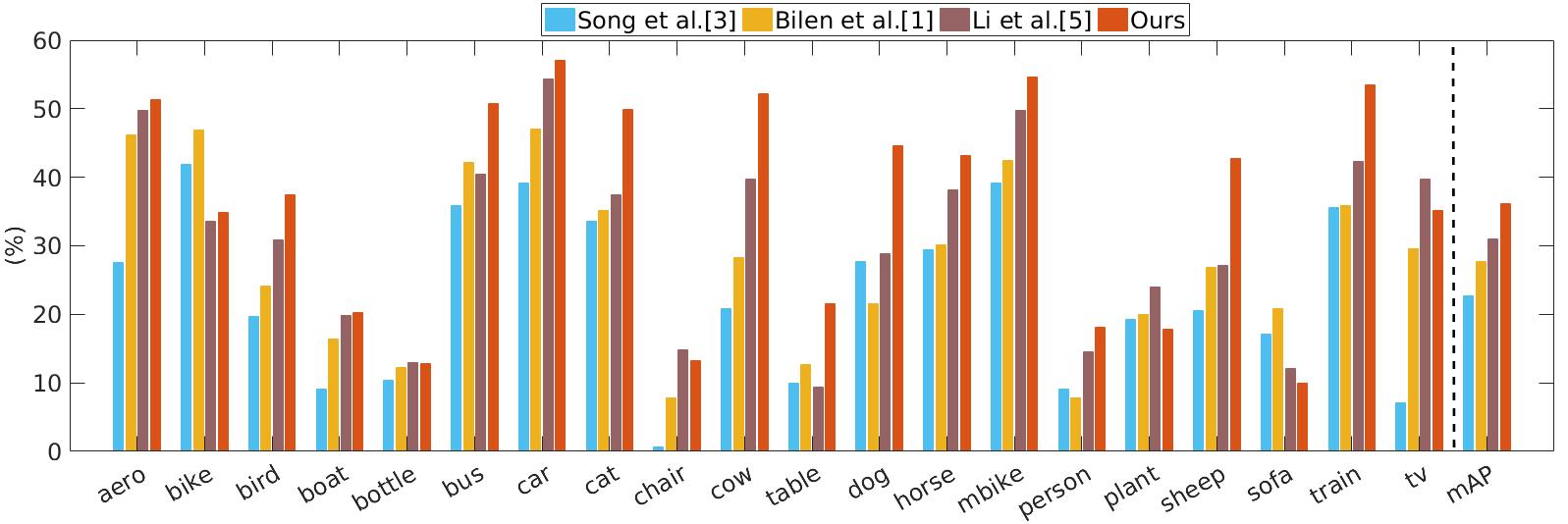}}
	\end{minipage}
	\caption{A comparison of our method (AlexNet) of detection mean average precision (mAP) on the PASCAL VOC 2007 dataset. Our method with the mAP (36.1\%) significantly outperforms other methods for most of the categories.}
	\label{fig_2}
\end{figure*}

\begin{table}[ht!]\setlength{\tabcolsep}{16pt}
	\caption{Quantitative comparison in terms of detection mean average precision (mAP) on the PASCAL VOC 2007 test set and correct localization (CorLoc) on the PASCAL VOC 2007 trainval set using AlexNet or VGGNet. The last rows show the mAP on the PASCAL VOC 2012 val set. We highlight the best performances and underline the 2nd best performances.}
	\label{table_1}
	\begin{center}
		\begin{tabular}{|l|cc|}
			\hline
			Methods~~~~~~~~~ (VOC 2007) & CorLoc&mAP \\
			\hline
			Li~ et~al. \cite{Authors14b}  &49.8&31.0 \\  
			Shi~ et~al. \cite{Authors14}~~~ (AlexNet) &\textbf{60.9}&\underline{36.0} \\  
			Our scheme &\underline{53.4}&\textbf{37.2}\\  
			\hline 
			Li~ et~al. \cite{Authors14b}  &52.4&39.5 \\ 
			Shi~ et~al. \cite{Authors14} &\textbf{64.7}&37.2 \\ 
			Bilen~ et~al. \cite{Alpher03}~~ (VGGNet)   &53.5&34.8  \\
			Jie~ et~al. \cite{jie2017deep} &56.1&40.8 \\
			Tang~ et~al. \cite{tang2017multiple} &\underline{60.6}&\textbf{41.2} \\
			Our scheme &55.9&\underline{40.9}\\
			\hline \hline
			Methods~~~~~~~~~~ (VOC 2012) & CorLoc&mAP \\
			\hline
			Li~ et~al. \cite{Authors14b}   &-&22.4\\  
			Our scheme~~~~~~ (AlexNet) &-&\textbf{25.3}\\
			\hline
			Li~ et~al. \cite{Authors14b}  &-&29.1\\  		
			Jie~ et~al. \cite{jie2017deep}  &54.8&\textbf{38.5} \\
			Tang~ et~al. \cite{tang2017multiple}~  (VGGNet) &\textbf{62.1}&\underline{37.9}  \\
			Our scheme&\underline{55.2}&35.2\\
			\hline
		\end{tabular}
	\end{center}
\end{table}

{\bf Evaluation metrics:}
To quantitatively evaluate the performance of the proposed method, we take two types of metrics, which are applied at the training and testing stage respectively. In the training dataset, we compute the percentage of images from which we obtain correct localization (CorLoc) \cite{deselaers2012weakly}. In the test dataset, we evaluate the performance of the object detector using mean average precision (mAP), a standard metric used in PASCAL VOC. Within both the metrics, we consider that a bounding box is correct if it has an IoU ratio of at least 50\% with the ground-truth object annotation.

{\bf Comparison with the state-of-the-art algorithms:}
We compared the proposed algorithm with the state-of-the-art methods dealing with the weakly supervised object localization problem \cite{song2014learning}. None of them use strong labels for training.

Fig. \ref{fig_2} shows the performance comparison between our proposed method developed with the AlexNet as baseline and the state-of-the-art WSL works \cite{Alpher02,Authors14b,song2014learning} on the VOC 2007 dataset. Models from Song \textit{et al.} \cite{song2014learning} and Bilen \textit{et al.} \cite{Alpher02} are MIL-based approaches with advanced model initialization. Our method is developed based on that from Li \textit{et al.} \cite{Authors14b}. Moreover, Tang \textit{et al.} \cite{tang2017multiple} propose an on-line instance classifier refinement, which classifies a fixed-size conv feature produced by some convolutional (conv) layers
with spatial pyramid pooling (SPP) layer. As the classifier is trained with the features from the SPP net, this model takes the advantage of a better initialization. In an entirely different way, we progressively adapt a classification network to an object detection network with denoised proposals as the pseudo strong labels.
Such domain adaptation helps to learn a better object detector from image-level annotated data. Unlike previous works
that rely on noisy proposals to localize the object candidates, we mine finer and class-specific proposals from the proposed work flow, which integrates the mask-out, MIL and subset proposal optimization. In addition, a fully model adaption is guaranteed with the re-training and re-weighting strategy.

\begin{figure}[ht!]
	\vskip -0.02in
	\begin{minipage}[b]{1.0\linewidth}
		\centering
		{\includegraphics[width=8cm]{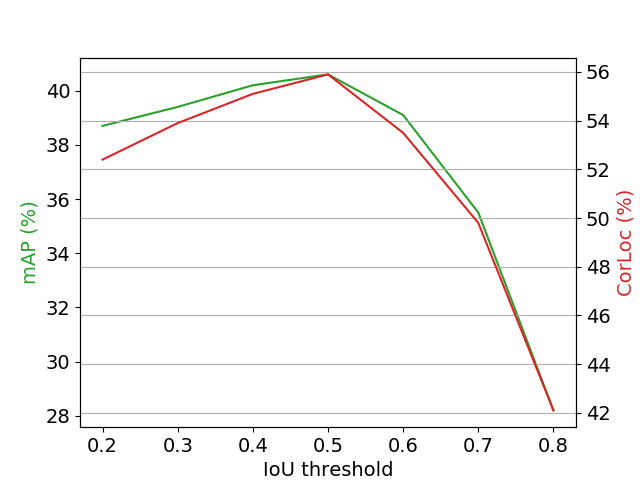}}
	\end{minipage}
	\caption{Performance over different IoU threshold of the VGG16 version on PASCAL VOC 2007}
	\label{fig_6}
\end{figure}

\begin{table}[ht!]\setlength{\tabcolsep}{10pt}
	\caption{Quantitative comparison in terms of detection mean average precision (mAP) on the PASCAL VOC 2007 test set for different re-training steps with AlexNet or VGGNet.}
	\label{table_2}
	\begin{center}
		\begin{tabular}{|l|ccccc|}
			\hline
			Re-train (mAP) & $0_{th}$ & $1_{st}$&$2_{nd}$&$3_{rd}$&$4_{rd}$ \\
			\hline
			AlexNet &31.0&34.1&36.8&37.2&37.2\\
			\hline
			VGGNet &38.5&39.9&40.3&40.9&40.9\\
			\hline
		\end{tabular}
	\end{center}
	
\end{table}

By incorporating the proposal subset optimization, the proposed model significantly outperforms other methods in terms of mAP for most of the categories. In Table \ref{table_1}, we make comparisons in terms of both the CorLoc and mAP on the training and testing set of the VOC 2007 dataset, respectively. In addition, we present the mAP on the val set of the VOC 2012. For other baseline methods, we list the best performances of the AlexNet and VGGNet models, which are reported in the paper. Based on the VGGNet, our method achieves 40.9\% mAP on VOC 2007 test set and 35.2\% mAP on VOC 2012 val set. It is also evident from Table \ref{table_1} that the detection performance is significantly improved by using a deeper network. Note that the method introduced by jie \textit{et al.} \cite{jie2017deep} is a regional CNN detector (Fast R-CNN \cite{girshick2015fast}). This model trained on seed samples is sufficiently powerful for selecting the most confident tight positives and is able to further train itself with the optimized proposals. We compare our method against this Fast RCNN based method by listing the results in Table \ref{table_1}. A similar performance is obtained by our model as the one on VOC 2007.

In addition to the standard IoU for evaluation, we analyze the influence over different IoU threshold in Fig \ref{fig_6}. It is evident that setting IoU = 0.5 achieves the best performance, and the results are not very sensitive to different values: when changing it from 0.5 to 0.6, the performance only drops a little bit. 

\begin{table}[ht!]\setlength{\tabcolsep}{6pt}
	\caption{Quantitative comparison in terms of computational time (hour) on the PASCAL VOC 2007 and 2012 training sets for different strategies.}
	\label{time}
	\begin{center}
	  \begin{tabular}{@{\extracolsep{4pt}}|l|cccccc|@{}}
			\hline
			\multirow{2}{*}{Strategy} & \multicolumn{2}{c}{BBox-initialization}&Re-weighting&\multicolumn{3}{c|}{Re-training} \\			\cline{2-3}\cline{4-4}\cline{5-7}
			 &Mask-out&MIL&Subset Optimization&AlexNet&VGG16&VGG19\\
			\hline
			VOC 2007&3&24&2&4&7&9\\
			\hline
			VOC 2012&7&36&3&7&14&17\\
			\hline
		\end{tabular}
	\end{center}
\end{table}

{\bf Impact of re-training strategy:}
The re-training strategy we utilized so far is straightforward. The process is to establish an order that adaptively optimize the refined proposals, and then fine-tune the detection network with the confident proposals. We notice that the proposals used to fine-tune the network are critical to train the baseline for detection. So it is promising to improve the annotation through an adaptive way. 

We use the same settings during the re-training stage as we adapt the classification network to a detection network. After training the detection network, we select the top 30 detection results and optimize them with the proposal subset optimization. Consequently, the training dataset is adaptively denoised and we obtain a better detection network. Table \ref{table_2} demonstrates that the mAP is increased from 31.0\% to 37.2\% for the AlexNet and from 38.5\% to 40.9\% for the VGGNet.

\begin{figure*}[t!]
	\begin{minipage}[b]{1.0\linewidth}
		\centering
		\centerline{\includegraphics[width=12cm]{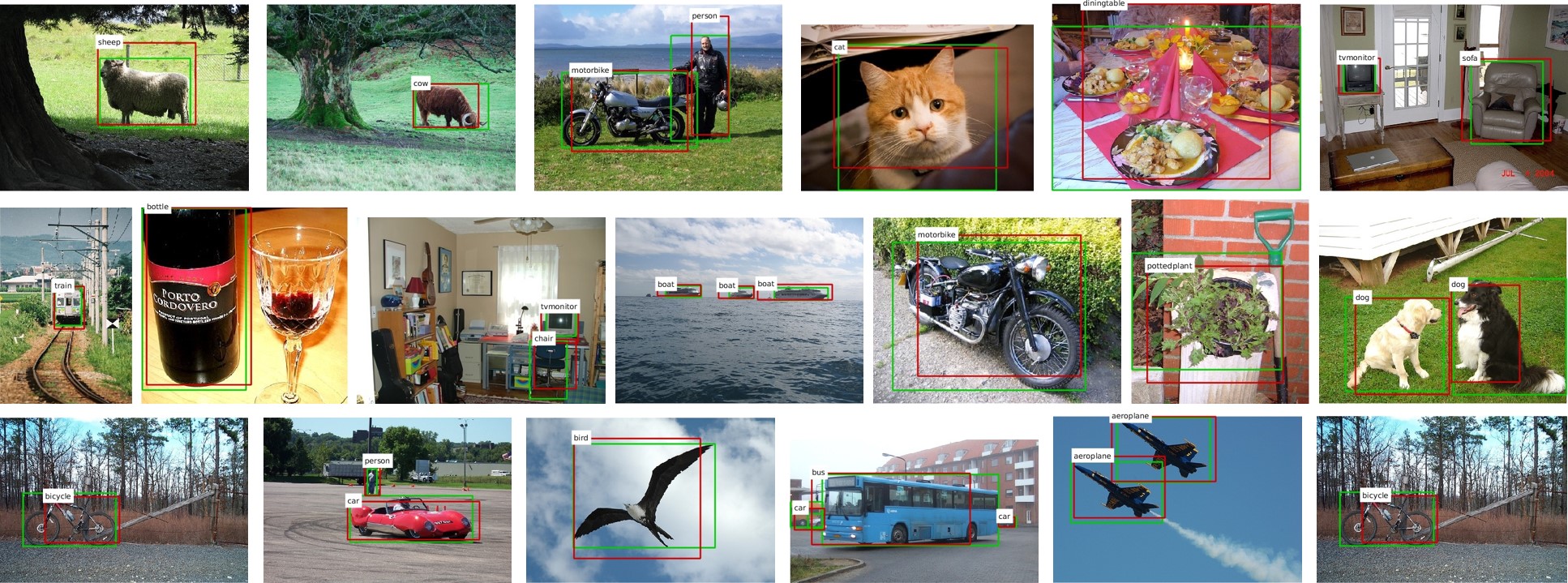}}
	\end{minipage}
	\caption{Sample detection results. Green boxes indicate ground-truth annotation. Red boxes indicate correct detections (with IoU $\geq$ 0.5). The sample images show the correct detections from different classes.}
	\label{fig_3}
\end{figure*}

\begin{figure}
	\vskip -0.02in
	\begin{minipage}[b]{1.0\linewidth}
		\centering
		{\includegraphics[width=8cm]{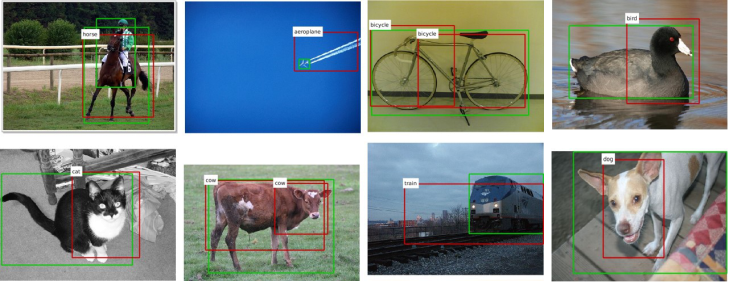}}
	\end{minipage}
	\caption{The sample images shows the wrong detections due to imprecise detection. Green boxes indicate ground-truth annotation. Red boxes indicate imprecise detections (with IoU $<$ 0.5). }
	\label{fig_4}
\end{figure}

{\bf Computational time analysis:}
We report the evaluation results on PASCAL VAL 2007 and PASCAL VAL 2012 in the paper. The re-training is conducted under AlexNet, VGG16, and VGG19. The training time of the experiments largely depends on the hardware resource. We train and evaluate the proposed method using the Intel Xeon(R) CPU E5-1607 v2 $@$ 3.00GHz $\times$ 4 and four K80 GPUs with 12 GB memory on a cluster. To reduce the training time of MIL, we employ 12 CPUs to separately train the MIL for each category of 21 classes. The training time of the experiment is shown in Table \ref{time}.

{\bf Error analysis:}
Fig. \ref{fig_3} shows some samples with accurate detections and Fig. \ref{fig_4} shows several examples with wrong detections. Our model often detects the correct objects in the image since we train the detector by incorporating a proposal subset optimization to improve the inaccuracy of the localization. Most of the model for WSL task may fail to predict a sufficient tight bounding box \cite{Authors14b}. The adaptive denoising part of Fig. \ref{fig_1} illustrates the procedure that the proposals are adaptively selected so that they gradually converge to the ground-truth of annotations. Nonetheless, the proposed model still has limitation as shown in the wrong detections in Fig. \ref{fig_4}. This is because our proposal subset optimization also depends on the detection scores even though it incorporates the overlaps of the proposals.

\section{Conclusion}
We have proposed a novel model by integrating adaptive proposal denoising strategies to handle the weakly supervised object localization problem. This approach first selects confident proposals by utilizing the output of the MIL framework as the starting point of training the detection network. At the training stage, we first adapt a pre-trained classification network with high confident proposals to a detection network, then re-weight the detection results with the proposal subset optimization method. The re-weighted proposals are taken to re-train the network, resulting a detection network that achieves competitive performance on PASCAL VOC datasets.  As a follow-up study, it is desire to adapt a new feature extraction method for the weakly supervised localization task. It is interesting to add the attention mechanism that assists to obtain attended features. We would like to introduce a module that effectively and efficiently extracts purified features.




\clearpage
\bibliographystyle{splncs04}
\bibliography{refs}
\end{document}